\documentclass{article}
\usepackage[preprint]{spconf,amsmath,graphicx}
\usepackage{tabularx,ragged2e}
\usepackage{subfig}
\usepackage{csquotes}
\usepackage{arydshln}

\DeclareMathOperator*{\argmax}{argmax}

\title{CROSS-MODAL KNOWLEDGE DISTILLATION FOR ACTION RECOGNITION}
\name{Authors Name(s)}
\address{Author Affiliation(s)}
\name{Fida Mohammad Thoker \hspace{1cm}Juergen Gall}
\address{
         University of Bonn, Germany \\
          \small{fmthoker@gmail.com \hspace{1cm} gall@iai.uni-bonn.de} 
          \sthanks{The work has been supported by the ERC Starting Grant ARCA (677650).} }
\begin{document}
%
\maketitle

\begin{abstract}

In this work, we address the problem how a network for action recognition that has been trained on a modality like RGB videos can be adapted to recognize actions for another modality like sequences of 3D human poses. To this end, we extract the knowledge of the trained teacher network for the source modality and transfer it to a small ensemble of student networks for the target modality. For the cross-modal knowledge distillation, we do not require any annotated data. Instead we use pairs of sequences of both modalities as supervision, which are straightforward to acquire. In contrast to previous works for knowledge distillation that use a KL-loss, we show that the cross-entropy loss together with mutual learning of a small ensemble of student networks performs better. In fact, the proposed approach for cross-modal knowledge distillation nearly achieves the accuracy of a student network trained with full supervision.  
\end{abstract}
\section{INTRODUCTION}
\label{sec:intro}

Action recognition is addressed in many works and in particular deep learning methods have been proposed for various modalities like RGB videos~\cite{DBLP:journals/corr/SimonyanZ14,TSN2016ECCV,CarreiraZ17,Fayyaz} or skeleton data \cite{hierarchical,stgcn2018aaai,DBLP:HCN,skeleton-based}. Deep learning methods for action recognition, however, require large annotated datasets. This poses a problem if the modality required for an application differs from the modality of an already annotated dataset. While acquiring data is usually not a bottleneck, annotating a dataset is very time consuming. It is therefore desirable to transfer the knowledge a network has learned from the already annotated dataset to a network for the new modality.   

For the cross-modal knowledge transfer, we assume that we have already trained a deep learning model for action recognition. This model is also called teacher network and we aim to distill~\cite{model_compression,44873,abs-1710-09282} and transfer the knowledge of the teacher network to the student network for the target modality. For the transfer, we assume that we have paired videos of both modalities that are not annotated. This assumption is not a constraint for most applications since acquiring videos with two different sensors at the same time is straightforward.        

In this work, we focus on the knowledge transfer from RGB videos to sequences of 3D skeleton poses \cite{Shahroudy_20z16_CVPR} since skeleton and RGB data are very different modalities in terms of data structure. To transfer the knowledge from the teacher network to the student network, we propose a different loss than the Kullback–Leibler (KL) divergence loss, which was used in~\cite{44873}. Instead of the KL-loss, we propose the cross-entropy for the transfer from the teacher to the student and train not one student network, but multiple student networks. Using an additional mutual loss for the student networks regularizes the transfer and increases the action recognition accuracy on the target modality.                     

We evaluate the approach on the NTU RGB+D dataset \cite{Shahroudy_20z16_CVPR} using ST-GCN \cite{stgcn2018aaai} and HCN \cite{DBLP:HCN} as network architectures for the student network. The experimental evaluation shows that the proposed approach outperforms an approach that uses the KL-loss as distillation and that it nearly achieves the accuracy of a student network trained with full supervision.

\section{RELATED WORKS}
\label{sec:related}
There is a large body of works on action recognition from 3D human pose data~\cite{YeZWZYG13}. More recently, most approaches use either recurrent neural networks to learn spatio-temporal features from sequences of skeleton data~\cite{song,HBRNN,Shahroudy_20z16_CVPR} or convolutional neural networks for classifying the skeleton sequences~\cite{kim,skeleton-based,new_representation}. In \cite{stgcn2018aaai}, a spatio-temporal graph convolutional network has been proposed to learn both spatial and temporal features directly from the skeleton data. A convolutional neural network is also used in \cite{DBLP:HCN} to learn co-occurrence features. It combines different levels of contextual information for learning co-occurrence features in a hierarchical manner. Both raw skeleton coordinates and their temporal differences are used within a two-stream framework.

Knowledge distillation has been originally proposed to compress ensemble classifiers into a smaller network without any significant loss of performance \cite{model_compression,44873}. In \cite{abs-1710-09282}, the approach has been extended to compress large networks and they showed that softening the softmax predictions of a network by a high temperature conveys important information, also called dark knowledge. Recently, knowledge distillation has been proposed for multi-modal action recognition. For instance, \cite{graph_dist} use a graph-based distillation method for action recognition that is able to distill information from multiple modalities during training. Similarly, \cite{multi3} proposed a multi-modal action recognition framework that uses multiple data modalities at training time. While these works analyze if the networks can be better trained using full supervision if additional modalities including the modality of the test data are available during training, we address the problem if the modality of the annotated training set differs from the modality of the test set. In \cite{Fayyaz}, a 3D convolutional neural network is initialized by transferring the knowledge of a pre-trained 2D CNN. Cross-modal distillation has been also used for other tasks such as object detection~\cite{gupta2015cross}, emotion recognition~\cite{emotion}, or human pose estimation~\cite{wall_pose}.

\section{CROSS-MODAL ACTION RECOGNITION}
\label{sec:method}

For cross-modal action recognition, we assume that a teacher network has been already trained on RGB videos. We now aim to train the student network for another modality, namely sequences of 3D human poses. For training the student network, we use pairs of RGB videos and human pose sequences. The pairs are not annotated and were therefore not part of the training data for the teacher network.           

\subsection{Teacher-Student Network}
\label{teach-stud}

\begin{figure}[t]

\begin{minipage}[b]{1.0\linewidth}
  \centering
  \centerline{\includegraphics[width=8.5cm]{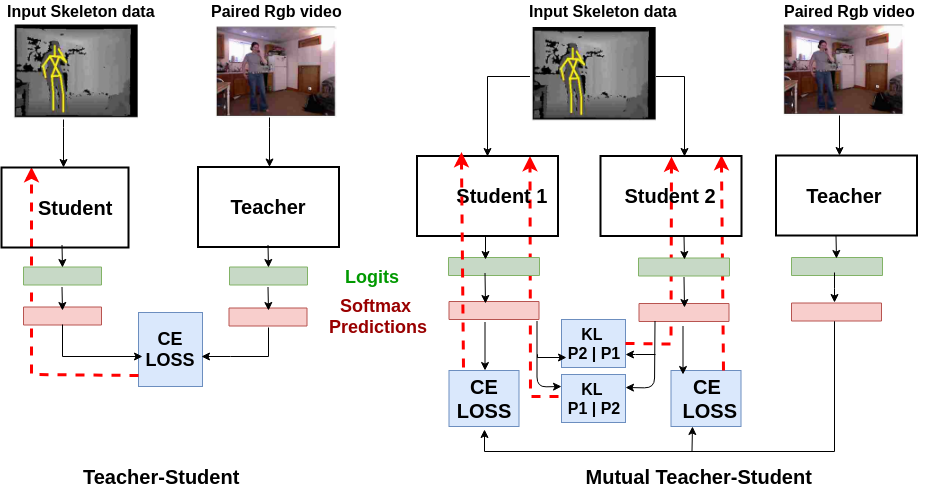}}
 
\end{minipage}
  \caption{
 \emph{(a)} The teacher network, which has been previously trained for RGB videos, provides the supervision for the student network for skeleton data. For training the student network, unlabeled pairs for both modality and the cross-entropy loss are used.  
  \emph{(b)} 
  Instead of one student network, two or more student networks can be trained together using mutual learning such that each student learns from the supervision of the teacher as well as the other student. The red dashed lines denote back-propagation for the corresponding loss functions.}
\label{fig:overview}
\end{figure}

The training of the student network is illustrated in Fig.~\ref{fig:overview}(a). The trained teacher network predicts for a training pair from the source modality the target class probabilities, where the vector of all class probabilities is denoted by $\mathrm{P_T}$. The parameters of the student network are then optimized such that the class probabilities $\mathrm{P_S}$ estimated by the student for the target modality matches $\mathrm{P_T}$. In~\cite{44873}, the Kullback–Leibler (KL) divergence has been proposed as loss for knowledge transfer between two networks of the same modality:   
\begin{equation}
\mathcal{KL}(\mathrm{P_S^\tau}, \mathrm{P_T^\tau})
=\sum_{c} {\mathrm{P_S^\tau}(c) \log \frac {\mathrm{P_S^\tau}(c)}  {\mathrm{P_T^\tau}(c)}}
\label{eq:losskl}
\end{equation}
where $\mathrm{P_S^\tau}$ and $\mathrm{P_T^\tau}$ are softmax predictions of the student and teacher networks both softened with temperature $\tau$: 
\begin{equation}
\mathrm{P^\tau}(c) = \frac{exp( \frac{z_c}{\tau})}
{\sum_{d} exp(\frac{z_d}{\tau})}.
 \label{eq:pred}
 \end{equation}
A temperature value of $\tau>1$ produces a softer probability distribution over the classes and has been proposed to avoid overfitting~\cite{44873}.


\subsubsection{Loss Function}
\label{ loss function } 

In our experimental evaluation, we show that the loss function \eqref{eq:pred} is not optimal for cross-modal knowledge transfer. In particular, finding an optimal $\tau$ is difficult and it strongly depends on the student network. Instead, we propose to use the cross-entropy loss      
  \begin{equation}
 \mathcal{CE}(\mathrm{P_S},\mathrm{P_T}) = -\log\left(\mathrm{P_S}( \hat{c}_T )\right)
\label{eq:hybridl}
\end{equation}
where $\hat{c}_T = \argmax_c \mathrm{P_T}(c)$. This means that the teacher makes a hard decision and we use the class label estimated by the teacher as supervision for the student network.     

\subsection{Mutual Learning}
\label{mutual}
In the context of fully supervised image classification, \cite{Zhang2017DeepML} proposed a deep mutual learning strategy. Instead of learning a single network with full supervision, an ensemble of networks is learned collaboratively and the networks teach each other throughout the training process.  

We show that mutual learning is also useful for cross-modal knowledge transfer. In this case, we train an ensemble of $K$ student networks together such that each network learns to mimic the probability distribution of the teacher network, as well as to match the probability estimates of its peers. Our approach for cross-modal knowledge transfer with mutual learning is shown in Fig.~\ref{fig:overview}(b) for $K=2$. 

Since the students are applied to the same modality, we can apply the KL-loss with softened temperature $\tau$ \eqref{eq:losskl}. The loss functions  $L_{\Theta_{1}}$ and $L_{\Theta_{2}}$ for the student networks with parameters  $\Theta_{1}$ and $\Theta_{2}$, respectively, are then given by 
  \begin{equation}
  \label{eq:dml1}
L_{\Theta_{1}} = \mathcal{CE}(\mathrm{P_1}, \mathrm{P_T}) + \mathcal{KL}(\mathrm{P_1^{\tau}}, \mathrm{P_2^{\tau}}) 
 \end{equation}
and
  \begin{equation}
  \label{eq:dml2}
L_{\Theta_{2}} = \mathcal{CE}(\mathrm{P_2}, \mathrm{P_T}) + \mathcal{KL}(\mathrm{P_2^{\tau}}, \mathrm{P_1^{\tau}}). 
 \end{equation}
The proposed approach can be extended to more student networks. For $K$ students, the loss function for optimizing the $k$-th student network is given by
  \begin{equation}
  \label{eq:DML}
L_{\Theta_{k}} = \mathcal{CE}(\mathrm{P_k}, \mathrm{P_T}) + \frac{1}{K-1}\sum_{l \neq k} \mathcal{KL}(\mathrm{P_k^{\tau}}, \mathrm{P_\textit{l}^{\tau}}).
 \end{equation}

\section{EXPERIMENTS}
\label{sec:expirement}
We evaluate our approach on the large scale multi-modal action recognition dataset NTU RGB+D~\cite{Shahroudy_20z16_CVPR}. 
The videos are collected from 40 distinct subjects and contain 60 different action classes. We use the RGB videos as source modality for the teacher network and the skeleton data as target modality. We adapt the cross subject evaluation protocol with 40,320 samples from 20 subjects for training and 16,560 samples from the remaining 20 subjects for testing. To evaluate the knowledge transfer, we divide the 20 training subjects into two groups of 10 subjects each, resulting in the \emph{Teacher-Train} set for training the teacher network and the \emph{Student-Train} set for training the student networks. While the RGB videos with class labels are used for the \emph{Teacher-Train} set, the \emph{Student-Train} comprises pairs of RGB videos and sequences of 3D human poses, but no class labels. We evaluate the accuracy of the student networks on the pose data of the \emph{Test} set.      
We use Temporal Segment Networks \cite{TSN2016ECCV} (TSN) as our teacher network and use optical flow as the teacher modality. We use the same hyper-parameters as in~\cite{TSN2016ECCV}. For the student networks, we use the Spatio Temporal Graph Convolution Networks (ST-GCN)~\cite{stgcn2018aaai} and the Hierarchical Co-occurrence Network (HCN)~\cite{DBLP:HCN} which both use the skeleton modality as their input data. We train the ST-GCN model using two GPUs with a batch size of 16 for a total of 200 epochs. All other hyper-parameters are the same as in \cite{stgcn2018aaai} and \cite{DBLP:HCN}.
  
   \begin{table}[t]
 \centering
 \begin{tabularx}{1.0 \linewidth}{Xccccccc}
 \hline
  Noise\% & 0 & 5 &10 & 14& 20& 25 \\ 
 \hline
  Acc &  78.50 & 73.20 & 72.58 & 71.51& 69.70 & 68.01 

 \end{tabularx}
 \caption{Impact of noisy labels during training on the classification accuracy of the ST-GCN model. The \emph{Student-Train} set is used for training and the \emph{Test} set for evaluation.}
 \label{table:noisy}
 \end{table}
 
In order to analyze how much knowledge we can extract from the teacher network, we evaluate the action recognition accuracy of the teacher network, which has been trained on the \emph{Teacher-Train} set. On the \emph{Student-Train} set, we obtain an accuracy of 86\%, i.e., the teacher network will produce around 14\% wrong labels during the knowledge transfer to the student networks. 

Next, we study the effect of noisy labels on the performance of the ST-GCN network. To conduct this experiment, we assign randomly wrong labels to a percentage of the training videos in the \emph{Student-Train} set. We then train the ST-GCN model on \emph{Student-Train} in a fully supervised manner using the noisy labels as ground-truth.  
Table \ref{table:noisy} reports the action recognition accuracy on the \emph{Test} set for different percentages of noisy labels during training. 78.5\% is the upper bound that can be achieved by cross-modal knowledge transfer if the teacher network is perfect since it corresponds to training the student network with full supervision. The accuracy drops from 78.5\% to 73.2\% if $5\%$ of the videos are wrongly labelled. Given that the teacher network misclassifies 14\% of the videos on \emph{Student-Train}, we can expect to achieve 71.51\% accuracy using cross-modal knowledge transfer.     

\begin{table}[t]
 \centering
 \begin{tabularx}{1.0 \linewidth}{Xcccccc}
 \hline
  $\tau$  & 1& 2& 5& 10& 20 &50 \\
 \hline
  Acc & 51.05 & 52.00 & 70.80 & 71.17 & 68.90 & 64.00 \\
 \hline
 \end{tabularx}
 \caption{Accuracy of the ST-GCN student network on the \emph{Test} set using the KL-loss with different values for the softmax temperature $\tau$.}
 \label{table:temp}
 \end{table}
 
Given some bounds for the accuracy that we can expect, we now analyse the impact of the loss functions for the task of cross-modal knowledge transfer. For the rest of the experiments, we train the student networks on \emph{Student-Train} using the teacher network as supervision and evaluate the action recognition accuracy of the student networks on the \emph{Test} set. We first analyze the impact of the temperature $\tau$ for the KL-loss~\eqref{eq:losskl}. Table~\ref{table:temp} shows that for $\tau\leq2$, the accuracy is very low since ST-GCN overfits on the \emph{Student-Train} set. 

\begin{table}[t]
 \centering
 \begin{tabularx}{1.0 \linewidth}{Xccc}
  \#Students &Method& Accuracy &Accuracy 	\\
  (K) & (supervision) & (Max) & (Average)\\
 \hline
 1 & Full Supervision & -	&	78.50	    \\
 1 & Teacher-Student & -	&	71.17	    \\
 2 & Ensemble without mutual  & 	71.93	&	72.32	    \\
 2 & Mutual Learning & 73.20	&	73.60	    \\
 3 & Mutual Learning & 73.60	&	74.22	    \\
 4 & Mutual Learning &  73.30 &	73.50		    \\
 \end{tabularx}
 \caption{Impact of mutual learning and the number of student networks $K$. In case of multiple student networks, we combine the predictions of the student networks during inference either by averaging the class probabilities or taking the maximum probability of each class. For the experiments, the KL-loss with $\tau=10$ is used.       
 } 
 \label{table:mutual}
 \end{table}
 
 We keep the KL-loss with $\tau=10$, but evaluate the benefit of using more than one student network for mutual learning. 
 Table \ref{table:mutual} reports the accuracy for mutual learning with multiple student networks (last three rows). In this case, we obtain an ensemble of student networks where the predictions are combined by averaging the class probabilities (average). We also report the results if for each class the highest probability among all student networks is taken (max). The results show that averaging performs better than taking the maximum. $K=3$ gives the best accuracy and mutual learning increases the accuracy compared to a teacher-student setup as proposed in~\cite{44873} by $+$3\%. To analyze if the improvement stems from the ensemble model or the mutual learning, we also trained two student networks without the mutual loss (ensemble without mutual). The result shows that 50\% of the improvement is due to the ensemble and the rest due to the mutual learning. It is interesting to note that mutual learning already achieves a higher accuracy than training the network with $5\%$ of randomly assigned wrong labels (Table \ref{table:noisy}).
 
   \begin{table}[t]
 \centering
 \begin{tabularx}{1.0\linewidth}{Xcc}
  Loss  & \# of students   &Accuracy 	\\
  \hline
 Full supervision	 & - &	78.50 \\
 
 
 KL & 1	&	71.17	    \\
 Cross-entropy  & 1 & 	74.91 \\
 KL + Mutual & 2	&	73.60	    \\
 Cross-entropy + Mutual   &2&  \bf{77.83}\\ 
 \end{tabularx}
 \caption{Results for the cross-entropy loss. For mutual learning, we average over the student networks. The cross-entropy loss outperforms the KL loss reported in Table~\ref{table:mutual}.
 } 
 \label{table:argmax_mutual}
 \end{table}

So far we have only used the KL-loss, but we have not evaluated the proposed approach using the cross-entropy loss~\eqref{eq:DML}. We report the results with the cross-entropy loss in Table~\ref{table:argmax_mutual}. Compared to the KL-loss, the accuracy increases from 71.17\% to 74.91\% for one student network and from 73.6\% to 77.83\% for two student networks. While the second term in \eqref{eq:DML} uses the KL-loss with $\tau=10$ for mutual learning, we observed that the accuracy decreases if cross-entropy is used for both terms. Compared to \cite{44873}, the proposed approach improves the accuracy by $+$6.66\%. Note that the proposed approach nearly achieves the accuracy of ST-GCN trained with full supervision.

\begin{table}[t]
 \centering
 \begin{tabularx}{1.0\linewidth}{Xcc}
  Loss  & \# of students   &Accuracy 	\\
  \hline
 Full supervision	 & - &	80.60 \\
 
 KL $(\tau=1)$  & 1&	74.40\\
 
  KL $(\tau=2)$   & 1	&	74.90\\

  Cross-entropy & 1& 	77.40 \\
  Cross-entropy + Mutual    &2&  79.00\\ 
 Cross-entropy + Mutual   &3&  \bf{79.50}\\ 

 \end{tabularx}
 \caption{Accuracy of the HCN student network on the \emph{Test} set using different loss functions and varying number of student networks.  
 } 
 \label{table:hcn}
 \end{table}

In order to demonstrate that the proposed approach is insensitive to the student network architecture, we also evaluated the accuracy of cross-modal knowledge transfer if we use HCN~\cite{DBLP:HCN} as student network. Table \ref{table:hcn} reports the results for the HCN model. For the KL-loss~\eqref{eq:losskl}, we had to adjust the temperature $\tau$. While ST-GCN performs better for a large value of $\tau$ as reported in Table~\ref{table:temp}, it is the other way around for HCN since HCN is a smaller network which suffers less from overfitting. For HCN, $\tau=2$ performs best and larger values of $\tau$ actually decrease the accuracy. This shows that it is very difficult to choose the hyper-parameter $\tau$~\cite{44873} in the context of cross-modal action recognition. If we use the proposed cross-entropy loss, this problem does not occur and it outperforms the KL-loss. If we use mutual learning with two or three students, the action recognition accuracy is improved by $+$4.1\% or $+$4.6\%, respectively, compared to KL with $\tau=2$~\cite{44873}. Note that we still use $\tau=10$ in \eqref{eq:DML} and we found that \eqref{eq:DML} is not sensitive to the parameter $\tau$ since the mutual loss is computed only for the student networks, which have the same network architecture applied to the same modality.                     

Finally, we compare our approach with the current state-of-the-art methods for the skeleton modality on the NTU RGB+D dataset in Table \ref{table:state-of-the-art}. Although our student networks are trained on less data and with less supervision, they achieve a higher accuracy than many other approaches that are trained with full supervision on the entire training set.

  \begin{table}[t]
 \centering
 \begin{tabularx}{1.0\linewidth}{Xcc}
  Method  & Full Train & \emph{Student-Train}	\\ 

 \hline
 Skeletal Quads \cite{skeletonquads}	&	38.62 & \\
 Lie Group \cite{ligroup}	&	50.08 &\\
 HBRNN-L \cite{HBRNN}	&	59.07 &\\ 
 Dynamic Skeletons \cite{ftpdynamic}	&	60.23 &\\
 PA-LSTM \cite{Shahroudy_20z16_CVPR}	&	62.90 &\\
 STA-LSTM \cite{song} & 73.40 &\\
 ST-LSTM+TS \cite{st-lstm-ts}	&	69.20 & \\
 Temporal Conv \cite{kim}	&	74.30 &\\
 VA-LSTM \cite{va-lstm}	&	79.20 &\\
 ST-GCN \cite{stgcn2018aaai}	&	81.60 & 78.50 \\
 Two-stream CNN \cite{skeleton-based} &	83.20&  \\
 HCN \cite{DBLP:HCN} &	86.50 & 80.60\\
 \hline
 \hline
 \emph{Cross-modal} ST-GCN  &&	77.83 \\
 \emph{Cross-modal}    HCN  &&	79.50 \\
 
 \end{tabularx}
 \caption{Comparison with the state-of-the-art for the cross-subject protocol. Note that the numbers are not directly comparable since the other approaches are trained with full supervision on the entire training set. While our approach is trained only on \emph{Student-Train} and with less supervision. 
 } 
 \label{table:state-of-the-art}
 \end{table}

%

 \vspace{-2mm}

\section{CONCLUSION}
\label{sec:conclusion}


We have presented an approach that uses knowledge distillation for cross-modal action recognition. The approach is able to transfer knowledge from one modality to another modality without the need of any additional annotations. Instead, pairs of sequences of both modalities are sufficient for the knowledge transfer. We evaluated our approach on a large-scale multi-modal dataset using two different student networks. In both cases, we showed that cross-modal knowledge transfer achieves an action recognition accuracy that is very close to fully supervised learning.

\vspace{-1mm} 

\bibliographystyle{main}
\bibliography{main}

\end{document}